%% file: main.tex
\documentclass[letterpaper, 10 pt, journal, twoside, final]{IEEEtran}

\input{macros.tex}
\begin{document}

\title{
CLIPPER+: A Fast Maximal Clique Algorithm \\ for Robust Global Registration }

\author{Kaveh Fathian$^{1}$, Tyler Summers$^{2}$
\thanks{Manuscript received: September 29, 2023; Revised December 28, 2023; Accepted January 15, 2024.}
\thanks{This paper was recommended for publication by
Editor Javier Civera upon evaluation of the Associate Editor and Reviewers’ comments.
This work is supported by the United States Air Force Office of Scientific Research under Grant FA9550-23-1-0424, by the National Science Foundation under Grant ECCS-2047040, and by the ARL SARA under Grant W911NF2420029.}
 \thanks{$^{1}$K.\ Fathian is with the Department of Computer Science, Colorado School of Mines. Email: fathian@ariarobotics.com}
 \thanks{$^{2}$T.\ Summers is with the Department of Mechanical Engineering, University of Texas at Dallas. Email: tyler.summers@utdallas.edu}
    \thanks{}
}%


\markboth{IEEE Robotics and Automation Letters. Preprint Version. Accepted January, 2024}
{Fathian \MakeLowercase{\textit{et al.}}: CLIPPER+} 

\maketitle


\begin{abstract}
We present CLIPPER+, an algorithm for finding maximal cliques in unweighted graphs for outlier-robust global registration. 
The registration problem can be formulated as a graph and solved by finding its maximum clique. This formulation leads to extreme robustness to outliers; however, finding the maximum clique is an NP-hard problem, and therefore approximation is required in practice for large-size problems. 
The performance of an approximation algorithm is evaluated by its
computational complexity (the lower the runtime, the better) and solution accuracy (how close the solution is to the maximum clique). 
Accordingly, the main contribution of CLIPPER+ is outperforming the state-of-the-art in accuracy while maintaining a relatively low runtime.
CLIPPER+ builds on prior work (CLIPPER \cite{lusk2021clipper} and PMC \cite{rossi2015parallel}) and prunes the graph by removing vertices that have a small core number and cannot be a part of the maximum clique. This will result in a smaller graph, on which the maximum clique can be estimated considerably faster. 
We evaluate the performance of CLIPPER+ on standard graph benchmarks, as well as synthetic and real-world point cloud registration problems. These evaluations demonstrate that CLIPPER+ has the highest accuracy and can register point clouds in scenarios where over $99\%$ of associations are outliers.
Our code and evaluation benchmarks are released
at \href{https://github.com/ariarobotics/clipperp}{\color{blue}https://github.com/ariarobotics/clipperp}.
\end{abstract}

\begin{IEEEkeywords}
Localization; SLAM; RGB-D Perception
\end{IEEEkeywords}

\section{Introduction}\label{sec:intro}

Data association is broadly defined as the correspondence of identical/similar elements across sets of data, and is a key component of many robotics and computer vision applications, such as localization and mapping~\cite{lajoie2020door,mangelson2018pairwise, ankenbauer2022view}, 
point cloud registration~\cite{bustos2017gore,yang2020teaser}, 
shape alignment~\cite{li2011robustly}, 
object detection~\cite{qi2017pointnet}, 
data fusion \cite{fathian2020clear, lusk2023mixer}, and 
multi-object tracking~\cite{rakai2022data}.
%
In these applications, it is crucial that data association is solved correctly and fast.

In point cloud registration, for example, we seek to find the rigid transformation (rotation/translation) that aligns two sets of 3D points.
This requires associating points in one set with their corresponding points in the other set.  
Local registration techniques such as the Iterative Closest Point (ICP) algorithm~\cite{besl1992method} associate points based on their nearest neighbor. These associations are generally wrong if the point clouds are not aligned well initially, leading to wrong registration. 
Better associations can be established by matching descriptors that are computed around each point in the point cloud and describe the local geometry and appearance of a point (e.g., classical FPFH \cite{rusu2009fast} or modern learning-based 3DMatch \cite{zeng20163dmatch}).
However, due to noise, repetitive patterns, small overlap between the point clouds, etc., these putative associations can have extreme outlier ratios (e.g., FPFH associations in Section~\ref{sec:pt_cld_reg} are $99\%$ outliers/wrong).
In these high-outlier regimes, existing outlier rejection techniques (such as the general RANSAC framework~\cite{fischler1981random} or specific frameworks for point cloud registration \cite{yang2020graduated}) either return wrong results or have impractical runtime (e.g., RANSAC's runtime grows exponentially in outlier ratio~\cite{bustos2017gore}).

To address these issues, we present the CLIPPER+ algorithm.
CLIPPER+ formulates the data association problem as a graph, in which the inlier/correct associations are the maximum clique. This formulation is robust to high outlier ratios and applicable to any problem that admits \textit{invariants} (see Section~\ref{sec:probform}) such as point, line, and plane could registration \cite{lusk2021clipper, lusk2022graffmatch, shi2021robin}.
To address the high computational complexity of finding the maximum clique (NP-hardness), CLIPPER+ finds an approximate solution instead, which is obtained from combining an improved version of our prior work, CLIPPER~\cite{lusk2021clipper}, and the greedy maximal clique algorithm in~\cite{rossi2015parallel}. 
CLIPPER+ runs in polynomial time and outperforms state-of-the-art algorithms in maximum clique estimation accuracy (Section~\ref{sec:dimcs}). 
Further, CLIPPER+ solutions are shown to be exact (i.e., the maximum clique) in over $99\%$ of the point cloud registration trials (Section~\ref{sec:pt_cld_reg}).

\begin{figure}[t]
\centering
\includegraphics[trim = 0mm 0mm 0mm 0mm, clip, width=0.90\columnwidth] {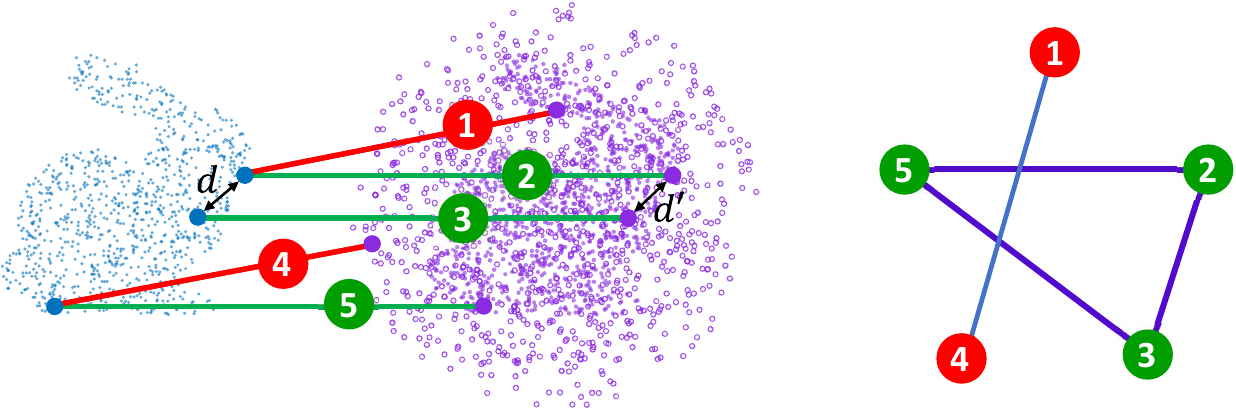}
\caption{An example of maximum clique formulation for robust global registration. (Left) Putative associations (red: outliers, green: inliers). (Right) Graph formulation with maximum clique indicating inlier associations.}
\vspace*{-0.5em}
\label{fig:opening}	
\end{figure}

\textbf{Contributions:}
In summary, this work's contributions are:
\begin{itemize}
    \item An improved solver (Algorithm~\ref{alg:relaxation}) leading to higher accuracy over our prior work CLIPPER~\cite{lusk2021clipper}.  
    \item The new CLIPPER+ algorithm as the combination of the greedy algorithm in \cite{rossi2015parallel} and CLIPPER~\cite{lusk2021clipper}, further improving both runtime and accuracy. 
    \item Evaluations demonstrating superior accuracy of CLIPPER+ over state-of-the-art on maximum clique problems, and correct registration of real-world point clouds in regimes with over $99\%$ outliers.
    \item Efficient C++ implementation of all algorithms (open-source code will be released).
\end{itemize}

\section{Related Work}\label{sec:}

\textbf{Robust registration:} 
In robotic applications of data association, such as registration, formulations that are robust to uncertainties (such as extreme noise and outliers) are based on globally optimal solvers, such as Branch and Bound  \cite{izatt2020globally, yang2015goicp, bustos2017gore}, and combinatorial approaches for maximum consensus \cite{chin2017maximum}. These methods can tolerate extreme outlier ratios, but have worst-case exponential complexity and are slow in practice. To address this issue, fast heuristics such as RANSAC~\cite{fischler1981random} and its variants \cite{barath2018graph}, 
graduated non-convexity \cite{yang2020graduated, black1996unification, blake1987visual}, or iterative local solvers for M-estimation \cite{agarwal2013robust, schonberger2016structure} are used, however, they are prone to failures in high-outlier regimes \cite{bustos2017gore,yang2020teaser, raguram2008comparative}.

\textbf{Graph-theoretic formulation:}
The dichotomy between robustness and runtime can be resolved in some data association settings that admit ``invariants'' (see Section~\ref{sec:probform}, and \cite{shi2021robin, lusk2021clipper, lusk2022graffmatch}) by formulation in a graph-theoretic framework. 
Recent algorithms that leverage this framework are 
CLIPPER \cite{lusk2021clipper}, TEASER~\cite{yang2020teaser}, and ROBIN \cite{shi2021robin} for global point cloud registration, robust to more than $99\%$ outliers and suitable for real-time applications, and
PCM~\cite{mangelson2018pairwise} for pairwise-consistent loop closures in SLAM.
This graph-theoretic formulation can be traced back to Bailey et al.~\cite{bailey2000data} for 2D LiDAR registration as a maximum common subgraph problem, for which the maximum clique indicated the correct data association.
Leordeanu and Hebert~\cite{leordeanu2005spectral} extended this graph-theoretic framework to weighted consistency graphs.
Enqvist et al.~\cite{enqvist2009optimal} noted the suboptimality of~\cite{leordeanu2005spectral} and proposed a vertex covering formulation, essentially an alternative to the maximum clique formulation of~\cite{bailey2000data}.
Parra et al.~\cite{bustos2019practical} proposed a practical maximum clique algorithm for geometric consistency based on branch and bound and graph coloring.
Recently, Zhang et al.~\cite{zhang20233d} have shown that the registration problem can be solved by relaxing the maximum clique constraint and using the maximal cliques instead.

\textbf{Maximum clique estimation:} 
Finding the maximum clique is a well-known \textbf{NP-hard} problem in its full generality \cite{wu2015review}, meaning that the complexity of finding the solution using exact algorithms 
(e.g., \cite{rossi2015parallel}) 
grows exponentially in graph size, which is impractical for data association applications.
An approximate (maximal clique) solution can however be found by polynomial-time algorithms.
Notable algorithms in this setting are Pelillo~\cite{pelillo1995relaxation} and Ding et al.~\cite{ding2008nonnegative}, 
based on the Motzkin-Straus formulation  \cite{motzkin1965maxima},  
Belachew and Gillis~\cite{belachew2017nmfmcp} based on symmetric rank-one nonnegative Matrix approximation, 
and our prior CLIPPER algorithm~\cite{lusk2021clipper} based on a continuous relaxation.
%

\section{Graph-Theoretic Robust Registration}\label{sec:probform}

The key idea for creating robustness to extreme outlier ratios is to find the \textit{largest set of jointly consistent} data and/or associations. 
This problem can be formulated as a graph, in which this set is represented by the maximum clique.
In what follows, we introduce this graph-theoretic framework and its use case for point cloud registration.

\textbf{Point cloud registration:}
The objective of point cloud registration is to find the rigid transformation that aligns a set of points to their corresponding points in another set.
The main challenge is finding the correct correspondences as usually only a subset of the points match, and, the points do not align perfectly due to noise.
An \textit{outlier} can be a point in one set that does not have a counterpart in the other set, or an association that matches wrong points across the sets. 
Our focus here is associations, and henceforth \textbf{outliers} imply \textit{outlier associations}.
Fig.~\ref{fig:opening} illustrates a point cloud registration example where we seek to find the blue bunny in the cluttered point cloud.

\textbf{Maximum likelihood solution:}
In the absence of prior knowledge and when outliers are random, unbiased, and unstructured, \textit{the largest set of jointly consistent associations} are inliers for the maximum likelihood solution.
For point cloud registration, two associations are \textbf{consistent} if their endpoints are equidistant, and therefore can be aligned by a rigid transformation.
In Fig~\ref{fig:opening}, the green associations align $3$ points while the red associations align $2$ points. 
Thus, the green associations are the inliers and can be used to compute the correct maximum likelihood solution.

\textbf{Graph formulation:}
Finding the largest set of jointly consistent associations can be formulated as a maximum clique problem. 
Given $n$ associations, the \textbf{consistency graph} is a graph of $n$ vertices where each vertex represents an association. 
An edge between two vertices indicates that the associations are consistent. In Fig.~\ref{fig:opening}, an edge between two vertices of the consistency graph (shown on the right) indicates consistency of corresponding associations. For example, there is an edge between vertices/associations $2$ and $3$ as their endpoints are equidistant ($d=d'$), while there is no edge between vertices/associations $1$ and $3$.
Given two associations with endpoint distances $d$ and $d'$, due to noise, often a \textbf{consistency threshold} $\epsilon$ is used where if $|d-d'|<\epsilon$, the associations are deemed consistent. 

A \textbf{clique} is a subset of vertices where every pair of vertices within that subset is connected by an edge.
The \textbf{maximum clique} is the clique with the largest number of vertices. A \textbf{maximal clique} is a clique that is not contained in a larger clique.
The maximum clique in Fig.~\ref{fig:opening} consists of vertices $2$, $3$, and $5$. Vertices $1$ and $4$ form a maximal clique.

\textbf{Invariants:} 
The graph-theoretic framework can be applied to a broad array of data association problems in robotics.
An \textbf{invariant} is a quantity that remains unchanged across the sets.
The invariant used for the point cloud registration example above was the Euclidean distance between the endpoints.
Invariants can be defined to register lines \cite{lusk2021clipper}, planes \cite{lusk2022graffmatch}, 2D-3D visual features \cite{shi2021robin}, etc.
Examples in this work, however, focus on point cloud registration.

\section{Maximal Clique Algorithms}\label{sec:algorithms}

\begin{algorithm}[t]
\caption{Degeneracy-Ordered Greedy Maximal Clique}
\label{alg:greedy}
\small
\begin{algorithmic}[1]
\State \textbf{Input} $A$: adjacency matrix; $K$: core numbers   
\State \textbf{Output} $C$: vertices that form maximal clique
\State $C \gets \{\}$; $c_{\max} \gets 0$ 
\State \algcomment{Sort vertices by core number in descending order:}
\State $(v_1, \dots, v_n) \gets$ sort vertices $v_i$ s.t. $K(v_i) \geq K(v_j)$ if $i < j$
\For {$i = 1 : n$}
    \If {$K(v_i) \geq c_{\max}$}
        \State \algcomment{Neighbors of $v_i$ with core numbers $\geq c_{\max}$:} 
        \State $S \gets \{v_j : A(v_i,v_j)=1, K(v_j)\geq c_{\max} \}$ 
        \State $(s_1, \dots, s_k) \gets$  sort $s_i\in S$ descending by core number
        \State $C' \gets \{\}$
        \For {$j=1:k$} \algcomment{For each vertex in sorted $S$}
            \If {$A(c_i,s_j)=1 ~\forall_{c_i \in C’}$}  \algcomment{$C’ \cup \{s_j\}$ is clique}
                \State $C’ \gets \{C’, s_j \}$ \algcomment{Add $s_j$ to $C’$}
            \EndIf
            \If {$|C’| > c_{\max}$} \algcomment{Found a larger clique}
                \State $C \gets C’$;  $c_{\max} \gets |C'|$ \algcomment{Update the output}
            \EndIf
        \EndFor
    \EndIf    
\EndFor
\end{algorithmic}
\vspace*{-0.3em}
\end{algorithm}

We present the main algorithmic contributions of this work in subsections~\ref{sec:algo_optim} and \ref{sec:clipperp}, where we discuss the maximal clique algorithms based on a continuous relaxation and CLIPPER+.
Subsection \ref{sec:algo_greedy} is a review of the approach in \cite{rossi2015parallel}, and is not an algorithmic contribution. However, our C++ implementation of this algorithm improved the performance and accuracy compared to its original implementation.

\subsection{Degeneracy-Ordered Greedy Maximal Clique Algorithm}\label{sec:algo_greedy}

Algorithm~\ref{alg:greedy} presents a greedy approach for finding a maximal clique.
Starting from an empty clique (line 3), we grow the clique one vertex at a time by looping through the vertices (line 6). For each vertex $v$ that this loop examines, the algorithm adds $v$ to the current clique if it is connected to every vertex that is already in the clique, and discards $v$ otherwise (lines 13-16). This algorithm has the overall runtime of $\O(\delta |E|)$, where $\delta$ is the maximum vertex degree and $|E|$ is the number of edges  \cite{rossi2015parallel}.

The maximal clique returned by the greedy approach depends on the initial vertex chosen to grow the clique, and the ordering of vertices (as they are sequentially examined to be added to the current clique or discarded).
A descending ordering of vertices by their core number greatly improves the odds of finding a large maximal clique (and possibly the maximum clique), as leveraged in Algorithm~\ref{alg:greedy} (lines 5 and 10).  
Mathematically, the \textbf{core number} or \textit{degeneracy} of a vertex $v$ is the largest integer $k$ such that the degree of $v$ remains non-zero when all vertices of degree less than $k$ are recursively removed from the graph.
The core number of vertices can be computed efficiently in $\O(|E|)$ 
by Batagelj and Zaversnik's algorithm \cite{batagelj2003m}.

\subsection{Continuous-Relaxation Maximal Clique Algorithm}\label{sec:algo_optim}

\begin{algorithm}[t]
\caption{Continuous-Relaxation Maximal Clique}
\label{alg:relaxation}
\small
\begin{algorithmic}[1]
\State \textbf{Input} $A$: adjacency matrix; $\bar{u}$: initial guess
\State \textbf{Output} $C$: vertices that form maximal clique 
\State \textbf{Parameters} $\sigma \gets 0.01$; $\beta \gets 0.5$; $ tol \gets 10^{-8}$ 
\State $M \gets A + I$  \algcomment {Add identity matrix $I$}
\State $\bar{M} \gets \mathbf{1} - M$ \algcomment{Binary complement of $M$}
\State $u\gets\max(\bar{u}/\|\bar{u}\|,0)$; ~ $d\gets d_0$;~  $\alpha \gets 1$
\While {$u$ not in binary state}
    \State $M_d \gets M - d \, \bar{M}$
    \State $F \gets u^\top M_d u$
    \State \algcomment{Gradient projected on $S^n$ tangent bundle:}
    \State $\nabla F_\perp(u) = 2 (I-uu^\top) M_d u$  
    \State $\Delta u \gets tol$; ~ $\Delta F \gets tol$
    \While {$\Delta u \nless tol$ or $\Delta F \nless tol$}
        \State $Armijo \gets \mathrm{False}$
        \While {$Armijo = \mathrm{False}$} \algcomment{Backtracking line search}
            \State $u_+ \gets u + \alpha \nabla F_\perp(u)$\ \algcomment{Solution update}
            \State $u_+ \gets \max(u_+/\|u_+\|,0)$ \algcomment{Retract to $\mathbb{R}^n_+\cap S^n$}
            \State $F_+ \gets u_+^\top M_d u_+$
            \State $\nabla F_\perp(u_+) = 2 (I-u_+ u_+^\top) M_d u_+$
            \State $\Delta F \gets F_+ - F$; ~ $\Delta u \gets u_+ - u$
            \State $Armijo \gets (\Delta F \geq \sigma \nabla F_\perp(u)^\top \Delta u )$
            \If {$Armijo = \mathrm{False}$}
                \State $\alpha \gets \alpha  \, \beta$ \algcomment{Reduce $\alpha$}
            \Else {~  $\alpha \gets \alpha / \sqrt{\beta}$} \algcomment{Increase $\alpha$}
            \EndIf            
        \EndWhile
        \State $u \gets u_+$; ~ $\nabla F_\perp(u) \gets \nabla F_\perp(u_+)$; ~ $F \gets F_+$
    \EndWhile
    \State $d\gets d + \Delta d$ \algcomment{Increase $d$}
\EndWhile
\State $C \gets \{i : u_i > 0\}$ \algcomment{Vertices that form maximal clique}
\end{algorithmic}
\vspace*{-0.3em}
\end{algorithm}

\textbf{Optimization formulation:}
The maximum clique problem in an undirected and unweighted graph of $n$ vertices can be formulated as
%
\begin{gather} \label{eq:mcp}
\begin{array}{ll}
\underset{u \in \{0,1\}^n}{\text{maximize}} & \sum_{i = 1}^{n}{u_i}
\\
\text{subject to} & u_i \, u_j = 0  \quad \text{if}~ A(i,j)=0, ~ \forall_{i,j}
\end{array}
\end{gather}
%
where $A \in \{0,1\}^{n\times n}$ is the \textbf{adjacency matrix} with $A(i,j)=1$ if and only if vertices $u_i$ and $u_j$ are connected. 
The optimization variable $u$ is a binary vector of $n$ elements, where $1$ entries indicate vertices that form a clique.
As $u$ is binary, the constraint $u_i \, u_j = 0$ if $A(i,j) = 0$  implies that if vertices $u_i$ or $u_j$ are disconnected, then at most one of them can be selected.
For example, consider the graph in Fig.~\ref{fig:opening} with the adjacency matrix $A$ and solution candidates $u, \bar{u}$ as
%
\begin{equation} \label{eq:numeric_example}
A =
\begin{bsmallmatrix}
0 & 0 & 0 & 1 & 0 \\
0 & 0 & 1 & 0 & 1 \\
0 & 1 & 0 & 0 & 1 \\
1 & 0 & 0 & 0 & 0 \\
0 & 1 & 1 & 0 & 0 \\
\end{bsmallmatrix}, 
~
u = 
\begin{bsmallmatrix}
0 \\
1 \\
1 \\
0 \\
1 \\
\end{bsmallmatrix},
~
\bar{u} = 
\begin{bsmallmatrix}
1 \\
0 \\
0 \\
1 \\
0 \\
\end{bsmallmatrix}.
\end{equation}
%
Both $u$ and $\bar{u}$ satisfy the constraints in \eqref{eq:mcp}. Solution $u$ is the global optimum with the objective value of $3$ (the size of the maximum clique, and the number of $1$ entries in $u$), while $\bar{u}$ gives the objective value of $2$.

By defining $M \eqdef A + I$, where $I$ is the identity matrix of appropriate size, it is straightforward to show that problem \eqref{eq:mcp} is equivalent to 
%
\begin{gather} \label{eq:belachew}
\begin{array}{ll}
\underset{u \in \{0,1\}^n}{\text{maximize}} & \dfrac{u^\top  M \, u}{u^\top u}
\\
\text{subject to} & u_i \, u_j = 0  \quad \text{if}~ M(i,j)=0, ~ \forall_{i,j}.
\end{array}
\end{gather}
%
This follows by observing that in \eqref{eq:mcp} and \eqref{eq:belachew} the constraints are identical in the sense that disconnected vertices cannot be selected jointly in the solution. Further, if $u^*$ is an optimal solution of \eqref{eq:belachew} and it has $m$ one entries, then the objective values of \eqref{eq:belachew} and \eqref{eq:mcp} will both be identical and equal to $m$ (e.g., $u$ and $\bar{u}$ in \eqref{eq:numeric_example} give objective values of $3$ and $2$ in \eqref{eq:belachew}).

\textbf{Continuous relaxation:}
To overcome the NP-hardness of problem \eqref{eq:belachew}, an approximate solution can be obtained by a continuous relaxation, where the binary domain ${u \in \{0,1\}^n}$ is relaxed to the set of non-negative real numbers ${u \in \br^n_+}$. 
Gradient-based optimization routines with polynomial time complexity can solve the relaxed problem, and the relaxed solution can be projected/rounded to the nearest binary.
The issue of this approach is that there is no guarantee that the binarized solution is a feasible solution that satisfies the constraints of the original problem \eqref{eq:belachew}.

Our proposed relaxation of \eqref{eq:belachew} that addresses this issue is 
%
\begin{gather} \label{eq:relaxation_1}
\begin{array}{ll}
\underset{u \in \br^n_+}{\text{maximize}} & \dfrac{u^\top  M_d \, u}{u^\top u},
\end{array}
\end{gather}
%
%
\begin{gather} \label{eq:Md}
M_d(i,j) \eqdef \left\{
\begin{array}{ll}
M(i,j) & \text{if} ~~ M(i,j) \neq 0 \\
-d & \text{if}  ~~ M(i,j) = 0
\end{array}
\right.
\end{gather}
%
where $d > 0$ is a positive scalar.
Problem \eqref{eq:relaxation_1} 
can be written equivalently as
%
\begin{gather} \label{eq:relaxation}
\begin{array}{ll}
\underset{u \in \br^n_+}{\text{maximize}} & F(u) \eqdef u^\top  M_d \, u 
\\
\text{subject to} & \| u \| = 1
\end{array}
\end{gather}
%
where $\| \cdot \|$ is the $\ell_2$ vector norm.\footnote{Vector $u$ can be written as $u = c u'$, where $c = \|u\|$, and $u' = u/c$ is a unit-norm vector. Replacing $u = c u'$ in \eqref{eq:relaxation_1} gives $\frac{u^\top M_d u}{u^\top u} = \frac{c^2 {u'}^\top M_d u'}{c^2 {u'}^\top u'} = {u'}^\top M_d \, u'$, demonstrating the equivalency to \eqref{eq:relaxation}.}

Our relaxation \eqref{eq:relaxation} is inspired by Belachew and Gillis~\cite{belachew2017nmfmcp}, which integrates the constraints in \eqref{eq:belachew} into \eqref{eq:relaxation} using the matrix $M_d$. 
The difference of this work with \cite{belachew2017nmfmcp} is that in \cite{belachew2017nmfmcp} the relaxation 
$\min_{u\in \br_+^n}\|M_d - u u^\top \|_F^2$ is used.
Intuitively, any entry $M_d(i,j) = -d$ penalizes joint selection of disconnected vertices $u_i$ and $u_j$ by the amount $-2 \,d$ in the objective.
Hence, as $d$ increases, the entries of $u$ that violate clique constraints converge to zero.

\textbf{Optimality guarantees:}
When $d\geq n$ (see \eqref{eq:Md}), the optima of the proposed relaxation  \eqref{eq:relaxation} are theoretically guaranteed to correspond to the optima of the original maximum clique problem \eqref{eq:belachew}.
That is, a local optimum of \eqref{eq:relaxation} corresponds to a maximal clique, and a global optimum of \eqref{eq:relaxation} corresponds to a maximum clique.
It is by no means trivial to prove this statement, and the interested reader should refer to Theorems~3-5 in \cite{belachew2017nmfmcp} that prove this statement for the relaxation $\min_{u\in \br_+^n}\|M_d - u u^\top \|_F^2$, and Theorem 2 in \cite{belachew2017nmfmcp} that establishes connections to our proposed relaxation \eqref{eq:relaxation}.

It is interesting to point out that any solution of \eqref{eq:relaxation} has a binary state (this follows from the analysis in \cite{belachew2017nmfmcp}). That is, the entries of a solution vector $u^* \in \br^n_+$ are either $0$ or equal to a positive scalar $c>0$. These positive entries are indicators of vertices that form a maximal clique. 
Noting that \eqref{eq:relaxation} can be (locally) solved in polynomial time by using a gradient-based solver, and the one-to-one correspondence between the optima of \eqref{eq:relaxation} and the (NP-hard) maximum clique problem \eqref{eq:belachew}, one may think that the maximum clique problem can be solved in polynomial time.
Note, however, that \eqref{eq:relaxation} is a nonlinear optimization problem, potentially with many local optima, and depending on the initial guess the solver can converge to a local optimum instead of the global optimum. Finding the global optimum of \eqref{eq:relaxation}, hence, remains an NP-hard problem.

\begin{algorithm}[t]
\caption{CLIPPER+}
\label{alg:clipperp}
\small
\begin{algorithmic}[1]
\State \textbf{Input} $A$: adjacency matrix
\State \textbf{Output} $C$: vertices that form maximal clique
\State $K \gets$ compute core number of vertices \algcomment{From  \cite{batagelj2003m}}
\State $C \gets$ Algorithm\ref{alg:greedy}($A,K$) \algcomment{Degeneracy-ordered greedy clique}
\State $\sI \gets \{i : K(v_i) \geq |C|\}$ \algcomment{Vertices with core number $\geq$ greedy clique size}
\State $A’ \gets A(\sI,\sI)$ \algcomment{Prune graph (only keep vertices in $\sI$)}
\If {$A’ = \emptyset$} {Terminate} \algcomment{Maximum clique found}
\EndIf
\State \algcomment{Binary complement of greedy clique in pruned graph:}
\State $\bar{u} \gets (\bar{u}_i : \bar{u}_i=0 ~\mathrm{if}~ i \in C, ~\mathrm{else},~ \bar{u}_i=1, ~ \forall_{i\in \sI})$
\State $C' \gets$ Algorithm\ref{alg:relaxation}($A',\bar{u}$) \algcomment{Continuous-relaxation clique}
\If {|C'| > |C|} {~ $C \gets C'$} \algcomment{Return larger clique}  
\EndIf
\end{algorithmic}
\vspace*{-0.3em}
\end{algorithm}

\begin{table*}[!t]
\scriptsize
\centering
\ra{1.1}
\caption{
Comparisons of the maximum clique estimation accuracy and runtime for CLIPPER+ on DIMACS benchmark \cite{johnson1996cliques} (bold is best).
}
\setlength{\tabcolsep}{4.8pt}
\begin{tabular}
{l c c c c c c c c c c c c c c c c c c c c c c c c}
\toprule
Benchmark & & & && \multicolumn{2}{c}{Pelillo} && \multicolumn{2}{c}{Ding} && \multicolumn{2}{c}{Belachew} && \multicolumn{2}{c}{CLIPPER} && \multicolumn{2}{c}{Greedy} && \multicolumn{2}{c}{Optim} && \multicolumn{2}{c}{\textbf{CLIPPER+}} \\ 
\cmidrule{1-4} \cmidrule{6-7} \cmidrule{9-10} \cmidrule{12-13} \cmidrule{15-16} \cmidrule{18-19} \cmidrule{21-22} \cmidrule{24-25}
Graph & $n$ & $s$ & $\omega_\mathrm{gt}$ && $t\downarrow$ & $r\uparrow$ && $t\downarrow$ & $r\uparrow$ && $t\downarrow$ & $r\uparrow$ && $t\downarrow$ & $r\uparrow$ && $t\downarrow$ & $r\uparrow$ && $t\downarrow$ & $r\uparrow$ && $t\downarrow$ & $r\uparrow$\\ 
\midrule
\verb|C125.9|         & 125 &  0.1 & 34 && 4.1 & 0.97 &&  4.5 & 0.56 && 10.9 &    \textbf{1} &&  9.1 &    \textbf{1} && \textbf{0.7} & 0.85 &&  0.8 &    \textbf{1} &&  1.6 &    \textbf{1} \\
\verb|C250.9|         & 250 &  0.1 & 44 && 4.7 & 0.89 &&  3.6 & 0.34 &&  7.8 & \textbf{0.95} &&  6.8 & \textbf{0.95} && \textbf{2.7} &  0.8 &&  6.4 & \textbf{0.95} &&  8.7 & \textbf{0.95} \\
\verb|brock200_2|     & 200 &  0.5 & 12 && 2.2 & 0.67 && 51.8 & 0.75 &&  4.3 & \textbf{0.83} &&    4 & \textbf{0.83} && \textbf{1.4} & \textbf{0.83} &&  2.6 & \textbf{0.83} &&  3.8 & \textbf{0.83} \\
\verb|brock200_4|     & 200 & 0.34 & 17 && \textbf{1.7} & 0.76 &&  9.4 & 0.88 &&  4.2 & 0.88 &&  8.9 & 0.65 && \textbf{1.7} & 0.82 && 29.4 & 0.82 &&  3.6 & \textbf{0.94} \\
\verb|gen200_p0.9_44| & 200 &  0.1 & 44 && 6.1 & 0.75 &&  3.9 &  0.2 && 10.4 & 0.84 && 10.9 & \textbf{0.89} && \textbf{2.2} & 0.73 &&  4.7 & \textbf{0.89} &&  6.4 & \textbf{0.89} \\
\verb|gen200_p0.9_55| & 200 &  0.1 & 55 && 4.6 & 0.69 &&  2.7 & 0.38 &&  4.9 & 0.69 &&  6.8 &    \textbf{1} &&   \textbf{2.0} & 0.64 &&  1.6 &    \textbf{1} &&  3.7 &    \textbf{1} \\
\verb|keller4|        & 171 & 0.35 & 11 && 2.3 & 0.64 &&    4.0 & 0.64 &&  2.8 & 0.73 &&  2.8 & 0.64 &&   \textbf{1.0} & \textbf{0.82} && 35.4 & 0.64 &&  5.8 & \textbf{0.82} \\
\verb|p_hat300-1|     & 300 & 0.76 &  8 && \textbf{1.3} & 0.75 && 33.9 & 0.88 &&  6.5 &    \textbf{1} &&  4.3 &    \textbf{1} && 2.1 & 0.88 &&  4.9 &    \textbf{1} && 10.7 &    \textbf{1} \\
\verb|p_hat300-2|     & 300 & 0.51 & 25 && \textbf{2.8} & 0.96 && 11.3 & 0.92 && 10.9 &    \textbf{1} && 15.9 & 0.96 && 3.7 & 0.84 &&  6.5 &    \textbf{1} && 10.1 &    \textbf{1} \\
\bottomrule
\end{tabular}
\\ 
\begin{flushleft}
{$n$: Number of graph vertices}.~ 
{$s$: Graph sparsity}.~
{$\omega_{\mathrm{gt}}$: Ground truth maximum clique size}. 
{$t$: Runtime (milliseconds); the lower, the better}.~
{$r$: Maximum clique accuracy ratio ($\hat{\omega}/\omega_{\mathrm{gt}}$); the closer to $1$, the better}.
\end{flushleft}
\label{tbl:dimacs}
\vspace*{-3.5em}
\end{table*}

\textbf{Optimization algorithm:}
We present a custom solver for \eqref{eq:relaxation} based on a projected gradient ascent approach, as described in Algorithm~\ref{alg:relaxation}. 
Our approach is similar to the algorithm in our previous work CLIPPER \cite{lusk2021clipper}; however, compared to \cite{lusk2021clipper}, Algorithm~\ref{alg:relaxation} is significantly improved and uses the Armijo procedure for selecting the appropriate step size, which leads to more accurate results (as we will show in the comparisons).

Problem \eqref{eq:relaxation} is nonlinear with many local optima in general (corresponding to maximal cliques). To improve the odds of finding the global optimum (maximum clique) and escape local optima, in Algorithm~\ref{alg:relaxation} we use a homotopy approach where we increase $d$ incrementally in an outerloop (lines 7-26).
As the penalty parameter $d$ increases incrementally by $\Delta d$ in each iteration of the outerloop (line 26), the elements of $u$ that violate the clique constraints are penalized further and $u$ converges to a feasible solution.
This process continues until $d$ is large enough ($d\geq n$) and $u$ converges to a binary state, corresponding to a maximal clique. 
The $\Delta d$ increments can be chosen as a small constant (as done in \cite{belachew2017nmfmcp}), however, in our implementation, we use a greedy scheme where we increase $d$ until the smallest element of $u$ that violates the clique constraint goes to zero in the next iteration. 
In the final step (line 27), the vertices of the maximal clique are identified as the non-zero elements of $u$.

The innerloop (lines 13-25) ensures that for each $d$ increment enough iterations of the gradient ascent are performed for the solution $u$ to reach a steady state.
Noting that the constraint manifold of the optimization problem \eqref{eq:relaxation} is $\mathbb{R}^n_+ \cap S^n$, where $S^n$ is the unit sphere, to speed up the convergence, we project the gradient $\nabla F(u) \eqdef 2 M_d \, u$ onto the tangent bundle of $S^n$ at $u$ and move along the orthogonal projection $\nabla F_\perp(u) = 2(I-u u^\top) M_d u$ (line 11).

To find an appropriate step size $\alpha$ along the projected gradient, we use backtracking line search (lines 15-24) with the Armijo procedure (lines 21-24), which guarantees a sufficient increase in the objective at each innerloop iteration. 
The convergence of the algorithm to a first-order optimal point is guaranteed by the convergence property of the projected gradient with Armijo steps~\cite{bertsekas1997nonlinear}. 
The solution update is computed in line 16, retracted back onto the constraint manifold (line 17), and the gradient ascent continues until convergence.

\textbf{Computational Complexity:} 
The worst-case complexity of Algorithm~\ref{alg:relaxation} is $\O(n^4)$. 
This is because the gradient computation (line 19) involves matrix-vector multiplications, which have $\O(n^2)$ complexity, and profiling shows this is where most of the time is spent.
The number of backtracking iterations (line 15) and gradient ascent iterations (line 13) can vary depending on the parameters and the data matrix, but it is linear in problem size ($\O(n)$) for quadratic objective $F(u) = u^\top M_d u$.
Lastly, the number of outerloop iterations (line 7) depends on $\Delta d$ increments (line 26) required to reach $d \geq n$ (at which point the solution is guaranteed to converge to the binary state). This is also linear in problem size.

\subsection{CLIPPER+ Maximal Clique Algorithm}\label{sec:clipperp}

Both Algorithms~\ref{alg:greedy} and \ref{alg:relaxation} are algorithms for finding maximal cliques.
The greedy approach of Algorithm~\ref{alg:greedy} runs fast but gives relatively less accurate estimates of the maximum clique size when the graph is not sparse.
In contrast, the optimization approach of Algorithm~\ref{alg:relaxation} is relatively slower but more accurate.
The main motivation of the proposed CLIPPER+ algorithm is to combine these two algorithms and thereby combine their relative benefits.

The enable this combination, recall that the \textbf{core number} of a vertex is the largest integer $k$ such that the degree of the vertex remains non-zero when all vertices of degree less than $k$ are removed. If a graph contains a clique of size $k$, then each vertex in the clique must have a degree of $k-1$ or larger, and therefore a core number of $k-1$ or larger.
For this reason, if we hope to find a larger clique, say of size $k+1$, then only vertices that have a core number $k$ or higher can be candidates. 
Using this observation, CLIPPER+, detailed in Algorithm~\ref{alg:clipperp}, first runs the greedy algorithm and obtains a maximal clique (line 4).
Assuming this clique has size $k$, if the graph contains a larger clique, then vertices must have core numbers of $k$ or larger. Therefore, the algorithm prunes the graph by removing vertices with core numbers strictly less than $k$ (line 6).
The pruning effectively limits the search space for the optimization (line 10) by reducing the number of vertices, which improves the runtime.

If the clique recovered by the greedy algorithm is the maximum clique, then
pruning the graph removes all the vertices and therefore we can terminate early (line 7).
This early termination particularly occurs when the graph is sparse, which leads to a significant speed-up over running the optimization-based algorithm on the original graph.

When the optimization-based algorithm is required, the initial guess used for the optimization can be chosen strategically to improve the chance of finding the maximum clique.
This can be done by choosing an initial guess vector that is a binary complement of the clique found from the greedy approach (line 9), and thus help the algorithm to converge to a different clique solution.
Lastly, the best solution is selected as the largest clique (line 11).

\textbf{Computational Complexity:} The worst-cast complexity of Algorithm~\ref{alg:clipperp} is $O(n^4)$. This is because it sequentially combines Algorithms~\ref{alg:greedy} and~\ref{alg:relaxation}, and thus inherits the highest worst-case complexity of its components (all other steps in Algorithm~\ref{alg:clipperp} have lower complexity). This basic worst-case complexity analysis is independent of any input graph structure. The numerical results in the experimental section provide a good indication of how the complexity translates to runtimes for various graph sizes in the practical applications.

\section{Experimental Evaluations}\label{sec:evaluations}

\begin{figure}[th!]
\centering
\includegraphics[trim = 0mm 0mm -5mm 0mm, clip, width=0.8\columnwidth] {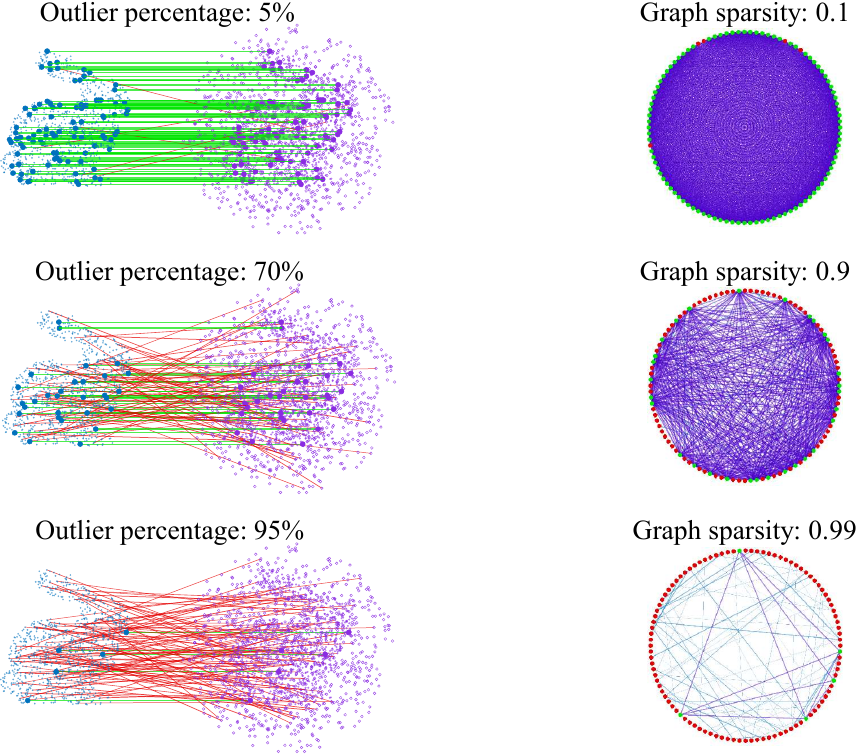}
\caption{The effect of outliers on the consistency graph. (Left) Putative associations  (red: outliers; green: inliers). (Right) Resulting consistency graph. The higher the outlier percentage, the sparser the graph.}
\vspace*{-0.4em}
\label{fig:bunny}	
\end{figure}

We evaluate the performance of CLIPPER+ (Algorithm~\ref{alg:clipperp}) in terms of the maximum clique estimation accuracy and runtime.
In addition, we provide ablation studies of CLIPPER+ by reporting the results of its standalone greedy (Algorithm~\ref{alg:greedy}) and optimization (Algorithm~\ref{alg:relaxation}) components. We show that CLIPPER+ achieves a performance superior to these standalone components through combining them.

\textbf{Algorithms:}
We test both classical and state-of-the-art maximum clique estimation algorithms. Classical works include
Pelillo \cite{pelillo1995relaxation} and Ding et al. \cite{ding2008nonnegative} based on relaxations of the Motzkin-Straus formulation \cite{motzkin1965maxima}.
State-of-the-art include our implementation of the greedy parallel maximum clique (PMC) algorithm~\cite{rossi2015parallel} in Algorithm~\ref{alg:greedy} (which is theoretically equivalent to ROBIN \cite{shi2021robin}), the algorithm by Belachew and Gillis~\cite{belachew2017nmfmcp}, and our prior CLIPPER algorithm
\cite{lusk2021clipper}.

\textbf{Benchmarks:}
Our evaluations examine finding the maximum clique on general graphs, as well as graphs that result from the graph formulation of synthetic/real-world point cloud registration problems.
While general graphs can have any structure, the registration graphs have certain patterns (e.g., sparsity) that significantly affect the results. Fortunately, as we will see, finding the maximum clique is empirically easier on graphs that result from registration.

\textbf{Platform and implementations:}
All benchmarks are run on a machine with an Intel Core i9 Processor and 32 GB RAM. The algorithms by Pelillo, Ding, and Belachew are implemented in Matlab by their original authors. Our prior work CLIPPER is also implemented in Matlab. 
The ``Greedy'' (Algorithm~\ref{alg:greedy}), ``Optim''  (Algorithm~\ref{alg:relaxation}), and CLIPPER+ algorithms are implemented in C++, which interface to Matlab via binary MEX binders for benchmarking.

\subsection{Maximum Clique Benchmark---DIMACS} 
\label{sec:dimcs}

\begin{figure}[t]
\centering
\includegraphics[trim = 0mm 0mm 15mm 0mm, clip, width=1.0\columnwidth] {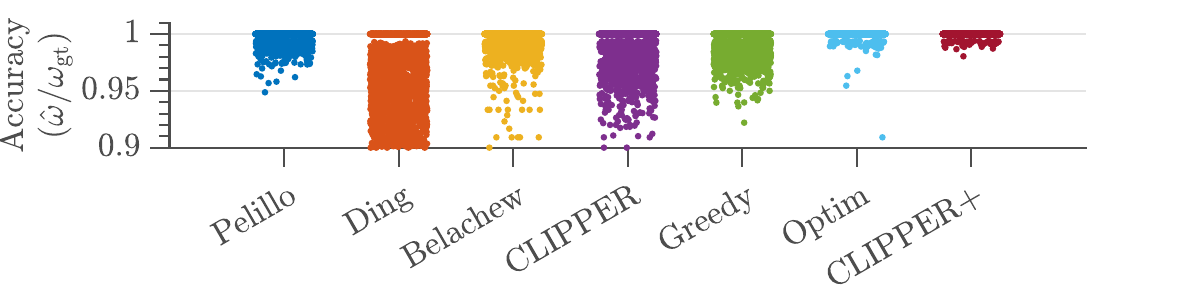}
\caption{The maximum clique estimation accuracy ratio on the Stanford Bunny benchmark (the closer to $1$, the better). Each point corresponds to the accuracy ratio from a single trial. CLIPPER+ outperforms all algorithms.}
\vspace*{-0.4em}
\label{fig:omega_ratio_box_bunny}	
\end{figure}

\textbf{Dataset:} 
To evaluate the algorithms on general graphs, we use the DIMACS benchmark \cite{johnson1996cliques}, which was introduced in 1996 and has been used widely since then to benchmark the maximum clique algorithms. 
The DIMACS dataset consists of graphs for which finding the maximum clique is challenging. In the interest of space, we present the result on a subset of smaller graphs in this dataset, shown in Table~\ref{tbl:dimacs}.
While DIMACS graphs are relatively small (100-4000 vertices), they are dense and contain more edges compared to graphs resulting from registration.
Despite the dataset being around for more than 25 years, the maximum clique of some graphs is still unknown, demonstrating the difficulty of the problem. 
For instance, on the \verb|C250.9| graph, the (multi-threaded) PMC exact algorithm \cite{rossi2015parallel}, used in this work for ground truth generation and in TEASER~\cite{yang2020teaser} for certifiable registration, took $28$ minutes on our machine to find the maximum clique.

\textbf{Evaluation:}
Table~\ref{tbl:dimacs} compares the accuracy and runtime of algorithms.
The \textbf{graph sparsity} is defined as $s \eqdef 1- \frac{|E|}{|E_{\max}|} \in [0,1]$, where $|E|$ is the number of graph edges, and $|E_{\max}|$ is the maximum possible number of edges. The small values of $s$ show that DIMACS graphs are generally dense.
The \textbf{accuracy ratio} is measured by $r\eqdef \hat{\omega}/\omega_{\mathrm{gt}}$, where $\hat{\omega}$ is the clique size found by the algorithm, and $\omega_{\mathrm{gt}}$ is the ground truth maximum clique size. The ratio of $1$ indicates that the maximum clique is found, hence, the closer $r$ is to $1$, the more accurate an algorithm is. 
\textit{CLIPPER+ outperforms all algorithms in the overall accuracy}. 
Unsurprisingly, the greedy algorithm has the best overall runtime due to its low computational complexity; however, it has a lower accuracy. 
By combining the greedy and optimization algorithms, CLIPPER+ obtains high accuracy and better overall runtime compared to the optimization-only approach (e.g., a $\sim$10x reduction of runtime from $29.4$ to $3.6$ milliseconds on \verb|brock200_2|).

\subsection{Registration Benchmark---Stanford Bunny}

\begin{figure}[t]
\centering
\includegraphics[trim = 0mm 0mm 7mm 0mm, clip, width=1.0\columnwidth] {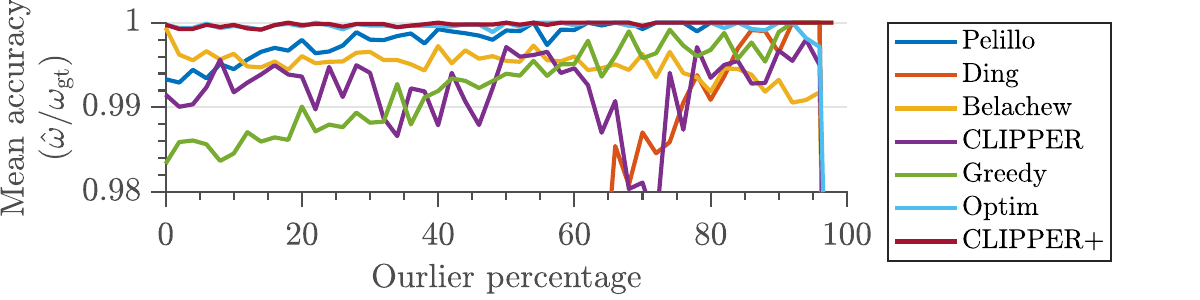}
\caption{Mean maximum clique estimation accuracy across different outlier percentages on the Stanford Bunny benchmark (the closer to $1$, the better).}
\vspace*{-0.2em}
\label{fig:omega_ratio_mean_bunny}	
\end{figure}

\begin{figure}[t]
\centering
\includegraphics[trim = 0mm 0mm 12mm 0mm, clip, width=0.95\columnwidth] {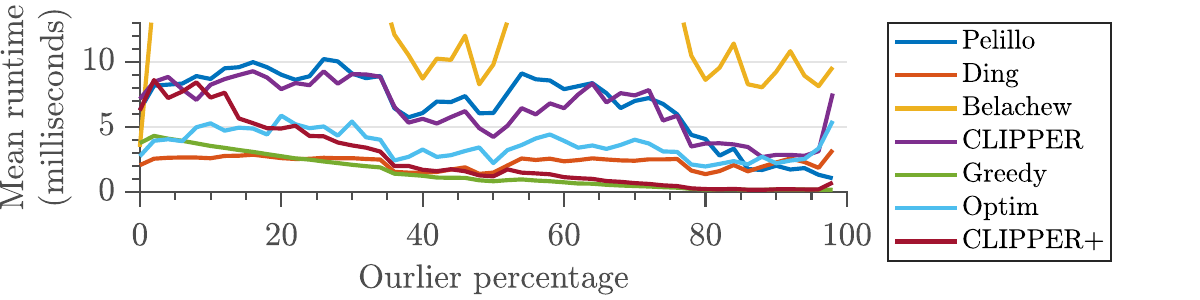}
\caption{Mean runtime across different outlier percentages on the Stanford Bunny benchmark (the lower, the better).}
\vspace*{-0.5em}
\label{fig:runtime_bunny}	
\end{figure}

\begin{figure*}[th!]
\centering
\includegraphics[trim = 0mm 0mm 0mm 0mm, clip, width=1.0\textwidth] {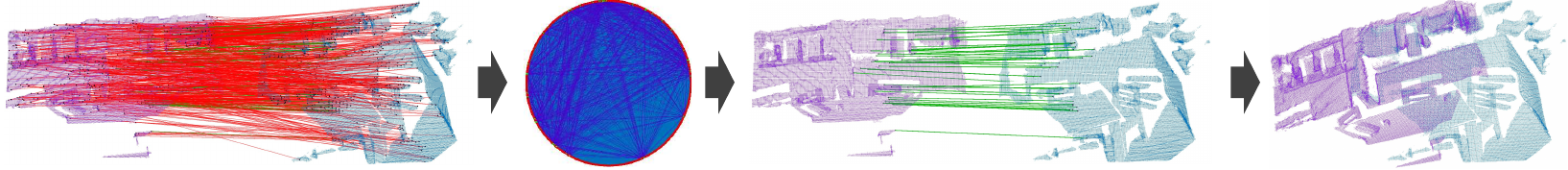}
\caption{A registration instance between two scans of the 7-Scenes dataset~\cite{shotton2013scene}. From left to right: $688$ putative FPFH associations having $93.6\%$ outliers (red lines); corresponding consistency graph used as input to CLIPPER+ (the maximum clique is highlighted); inlier associations corresponding to the maximum clique solution found by CLIPPER+; aligned point clouds using the rotation/translation computed from the Arun's method \cite{arun1987least} using inliers.}
\vspace*{-0.3em}
\label{fig:registration_demo}	
\end{figure*}

\begin{table*}[!t]
\scriptsize
\centering
\ra{1.1}
\caption{
Comparisons of maximum clique estimation accuracy and runtime of CLIPPER+ on real-world point cloud registration datasets (bold is best).
}
\setlength{\tabcolsep}{2.2pt}
\begin{tabular}
{l l c c c c c c c c c c c c c c c c c c c c c c c c c}
\toprule
Benchmark & & & & & && \multicolumn{2}{c}{Pelillo} && \multicolumn{2}{c}{Ding} && \multicolumn{2}{c}{Belachew} && \multicolumn{2}{c}{CLIPPER} && \multicolumn{2}{c}{Greedy} && \multicolumn{2}{c}{Optim} && \multicolumn{2}{c}{\textbf{CLIPPER+}} \\ 
\cmidrule{1-6} \cmidrule{8-9} \cmidrule{11-12} \cmidrule{14-15} \cmidrule{17-18} \cmidrule{20-21} \cmidrule{23-24} \cmidrule{26-27}
Name & Sequence & $N$ & $\bar{op}$ & $\bar{n}$ & $\bar{s}$ && $\bar{t}\downarrow$ & $\bar{r}\uparrow$ && $\bar{t}\downarrow$ & $\bar{r}\uparrow$ && $\bar{t}\downarrow$ & $\bar{r}\uparrow$ && $\bar{t}\downarrow$ & $\bar{r}\uparrow$ && $\bar{t}\downarrow$ & $\bar{r}\uparrow$ && $\bar{t}\downarrow$ & $\bar{r}\uparrow$ && $\bar{t}\downarrow$ & $\bar{r}\uparrow$\\ 
\midrule
7-Scenes 
 & \verb|redkitchen| & 244 & 92.8\% & 680 & 0.88 && 118.5 & 0.96 && 28.3 & 0.9 && 73.7 & \textbf{0.99} && 62.7 & 0.97 && \textbf{10.1} & 0.89 && 119.9 & \textbf{0.99} && 53.1 & \textbf{0.99} \\
\cmidrule{8-27}
\multirow{3}{*}{Sun3D}
 & \verb|home_at_scan1| & 81 & 88\% & 670 & 0.83 && 133.1 & 0.97 && 17.5 & 0.87 &&   64.0  & 0.98 && 58.2  & 0.98 && \textbf{10.3} & 0.89 &&  76.0   & \textbf{0.99} && 46.4 & \textbf{0.99} \\
 & \verb|hotel_uc_scan3| & 117 & 93.5\% & 843 & 0.89 && 215.3 & 0.96 && 28.9 & 0.89 && 116.5 & \textbf{0.99} && 109.3 & 0.98 && \textbf{12.3} & 0.89 && 159.4 & \textbf{0.99} && 56.1 & \textbf{0.99} \\
 & \verb|mit_76_1studyroom2| & 96 & 94.2\% & 936 & 0.9 && 192.9 & 0.97 && 32.2 & 0.87 && 120.6 & \textbf{0.99} && 102.8 & 0.97 &&  \textbf{14}  & 0.88 && 180.8 & \textbf{0.99} && 83.9 & \textbf{0.99} \\
\cmidrule{8-27} 
\multirow{2}{*}{ETH}
 & \verb|gazebo_summer| & 116 & 98.3\% & 4320 & 0.97 && 4255.2 & 0.95 && 1275.6 & 0.8  && 8467.8 & 0.98 && 7978.5 & 0.97 && \textbf{351.5 }& 0.81 && 7595.7 & \textbf{0.99} &&  3769 & \textbf{0.99} \\ 
 & \verb|wood_autumn| & 87 & 99.4\% & 5703 & 0.98 && 4866.2 & 0.94 && 2990.9 & 0.89 && 14464  & 0.98 &&  12774 & 0.97 && \textbf{564.5} & 0.75 &&  14758 & \textbf{0.99} && 14594 & \textbf{0.99} \\ 
\bottomrule
\end{tabular}
\\ 
\begin{flushleft}
{$N$: Total number of overlapping point cloud scans on which registration is performed}.~
{$\bar{op}$: Mean of outlier percentages in putative associations}.~
{$\bar{n}$: Mean graph size}.~
{$\bar{s}$: Mean graph sparsity}.~
{$\bar{t}$: Mean runtime (milliseconds); the lower, the better}.~
{$\bar{r}$: Mean maximum-clique accuracy ratio ($\hat{\omega}/\omega_{\mathrm{gt}}$); the closer to $1$, the better}.
\end{flushleft}
\label{tbl:registration}
\vspace*{-3.5em}
\end{table*}

Using the Stanford Bunny point cloud~\cite{curless1996stanfordbunny}, shown in Fig.~\ref{fig:bunny}, we evaluate CLIPPER+ as an algorithm for global point cloud registration in various outlier regimes.

\textbf{Dataset:}
The (downsampled) Bunny point cloud consists of $1000$ points that fit in a cube of size $0.2 \,\mathrm{m}$.
To generate a second point cloud, we add uniform noise in the range $[-\epsilon/2, \epsilon/2]$ to all points, where $\epsilon$ is set as the mean distance of all points to their nearest neighbors in the Bunny point cloud.
Additionally, $1000$ outlier points randomly drawn from a sphere of radius $1 \,\mathrm{m}$, centered at the bunny point cloud, are added to simulate clutter.
From the set of all possible associations between the points in the two point clouds, $200$ associations are randomly selected from the set of inlier and outlier associations (we keep this number small to be able to find the ground truth maximum clique).
We consider different outlier ratios (ranging from $0\%$ to $98\%$ in $2\%$ increments), to test scenarios with various data association accuracy, as shown in Fig.~\ref{fig:bunny}.
%
%
We use the noise $\epsilon$ as the threshold to generate the consistency graph (according to Section~\ref{sec:probform}). 
The ground truth maximum clique is found by running the exact PMC algorithm in \cite{rossi2015parallel}. 
Evaluations of maximum clique estimation accuracy and runtime are performed across $50$ Monte Carlo runs/graphs for each increment of the outlier percentage.

\textbf{Evaluation:}
Fig.~\ref{fig:omega_ratio_box_bunny} compares the accuracy ratio $r$ of algorithms across all runs and all outlier ratios. 
\textit{CLIPPER+ clearly demonstrates the highest accuracy} because the distribution of $r$ is closest to $1$. 
Interestingly, while the greedy and optimization algorithms returned low-accuracy solutions in some instances, by combining these algorithms CLIPPER+ obtains an accuracy beyond these standalone components. 

Fig.~\ref{fig:omega_ratio_mean_bunny} shows the mean of the accuracy ratio $r$ (averaged across $50$ Monte Carlo runs at each outlier percentage increment) versus the outlier percentage.
The accuracy of the greedy approach is low in low-outlier regimes (which have dense consistency graphs, see Fig.~\ref{fig:bunny}). As the outlier percentage increases and the graph becomes sparser, the accuracy of the greedy method improves.
Both CLIPPER+ and optimization algorithms retain a high accuracy ratio across all outlier percentages.

The mean runtime of the algorithms is shown in Fig.~\ref{fig:runtime_bunny}. 
The runtime of the optimization method remains roughly the same, while the runtime of the greedy method improves as the outlier ratio grows and the graph becomes sparser.
The runtime of CLIPPER+ in low outlier regimes is roughly equal to the compound runtimes of its greedy and optimization components because in such regimes pruning the graph based on core numbers does not remove a significant number of vertices (if any). However, as the outlier ratio increases and the graph becomes sparser, pruning removes more vertices and its speed-up effect becomes more apparent. This can be seen around the $20\%$ outlier ratio, where CLIPPER+ becomes faster than the optimization method, and its speed improves further as the outlier percentage increases.

\subsection{Registration Benchmark---Real-World Point Clouds}
\label{sec:pt_cld_reg}

\textbf{Datasets:}
We use sequences in the real-world 7-Scenes~\cite{shotton2013scene},
Sun3D~\cite{xiao2013sun3d}, and ETH \cite{pomerleau2012challenging} datasets (similar to 3DMatch \cite{zeng20163dmatch} and 3DSmoothNet \cite{gojcic20193DSmoothNet} evaluations). 
Sun3D and 7-Scenes are dense indoor RGB-D point clouds, while ETH is an outdoor LiDAR point cloud.
In each sequence, we consider pairs of point clouds (or scans) that have an overlap. To increase registration speed, we downsample the point clouds
by discretizing the 3D space into cubes of size $\epsilon = 0.05 \, \mathrm{m}$ for 7-Scenes and Sun3D datasets, and $\epsilon = 0.1 \, \mathrm{m}$ for the ETH dataset, and using the mean of the point coordinates in each cube as a single-point representative.
For each downsampled point, FPFH descriptor vectors \cite{rusu2009fast} are computed and associated bilaterally based on their $l_2$ norm distance using the k-nearest neighbors algorithm. 
To generate the ground truth for our evaluations, we use the exact maximum clique algorithm of \cite{rossi2015parallel} to find the largest set of geometrically consistent associations in these putative FPFH associations.
We store results if this maximum clique solution correctly registers the point clouds according to the ground truth provided by the datasets---the maximum clique/likelihood solution may register points wrongly due to practical limitations such as repetitive patterns (perceptual aliasing), insignificant overlap between the point clouds, and lack of any inlier associations caused by downsampling and FPFH inaccuracies.
For evaluations, the consistency graph is generated according to Section~\ref{sec:probform} from putative FPFH associations, using the downsampling $\epsilon$ as the consistency threshold. An evaluation instance is shown in Fig.~\ref{fig:registration_demo}.

\textbf{Evaluation:}
Table~\ref{tbl:registration} presents the evaluation results for all datasets and algorithms. 
\textit{CLIPPER+ outperforms all algorithms in accuracy on all datasets/sequences.}
The greedy algorithm has the smallest runtime at the expense of the lowest overall accuracy.
The standalone optimization algorithm has similar accuracy to CLIPPER+. However, except on the last sequence, it is around $2$x slower. 
This demonstrates the advantage of CLIPPER+ over its standalone greedy and optimization components.

Lastly, we point out the high outlier ratios of FPFH associations (e.g., on average, $99.4\%$ of associations in the \verb|wood_autumn| sequence are outliers).
Due to the maximum clique solution correctly registering the point clouds in our benchmark, the accuracy ratio $r$ shows the point cloud registration success rate.
Thus, \textit{CLIPPER+ correctly registers the point clouds in $99\%$ of the trials despite extreme outlier percentages.}
This is what distinguishes CLIPPER+ from existing robust registration frameworks (such as RANSAC~\cite{fischler1981random}), that can fail in these high-outlier regimes.

\section{Conclusion and Future Directions}\label{sec:conclusion}

We presented CLIPPER+, a maximal-clique-finding algorithm for unweighted graphs that enables robust global registration in robotics and computer vision applications. Future work includes investigating alternative optimization methods such as second-order or quasi-Newton methods, and an extension to the weighted graphs based on our previous work \cite{lusk2021clipper} (designed for weighted graphs) and an extension of core numbers to weighted settings (studied in \cite{liu2020incremental}).  We also plan to integrate the algorithm in point cloud registration pipelines \cite{yang2020teaser} as the outlier rejection module to improve the runtime and outlier rejection capacity.


\balance 

\bibliographystyle{IEEEtran}
\bibliography{refs}

\end{document}

%% file: macros.tex
\usepackage{amsmath} 
\usepackage{amssymb} 
\usepackage{amsfonts}
\usepackage[dvipsnames]{xcolor}
\usepackage{graphicx}
\usepackage{tabularx}
\usepackage{multirow}
\usepackage{mathtools}
\usepackage{esvect} 
\usepackage[binary-units=true]{siunitx}
\usepackage{caption}
\usepackage[pdftex, pdfstartview={FitV}, pdfpagelayout={TwoColumnLeft},bookmarksopen=true,plainpages = false, colorlinks=true, linkcolor=black, citecolor = black, urlcolor = black,filecolor=black , pagebackref=false,hypertexnames=false, plainpages=false, pdfpagelabels ]{hyperref}
\usepackage[T1]{fontenc} 
\usepackage{mathtools, cuted} 

\usepackage{balance} 
\usepackage{booktabs} 
\usepackage[export]{adjustbox} 

\newcolumntype{x}[1]{%
>{\raggedleft\hspace{0pt}}p{#1}}%

\usepackage{algorithm}
\usepackage[noend]{algpseudocode} 
\definecolor{commentclr}{RGB}{34, 139, 34}
\newcommand{\algcomment}[1]{{\color{commentclr}\% #1}}

\usepackage{tikz}

\usepackage{subcaption}
\DeclareCaptionLabelSeparator{periodspace}{.\quad}
\newcommand{\ra}[1]{\renewcommand{\arraystretch}{#1}}

\usepackage{amsthm}

\theoremstyle{definition}

\usepackage{letltxmacro}
\LetLtxMacro\orgvdots\vdots
\LetLtxMacro\orgddots\ddots

\makeatletter
\DeclareRobustCommand\vdots{%
	\mathpalette\@vdots{}%
}
\newcommand*{\@vdots}[2]{%
	\sbox0{$#1\cdotp\cdotp\cdotp\m@th$}%
	\sbox2{$#1.\m@th$}%
	\vbox{%
		\dimen@=\wd0 %
		\advance\dimen@ -3\ht2 %
		\kern.5\dimen@
		\dimen@=\wd2 %
		\advance\dimen@ -\ht2 %
		\dimen2=\wd0 %
		\advance\dimen2 -\dimen@
		\vbox to \dimen2{%
			\offinterlineskip
			\copy2 \vfill\copy2 \vfill\copy2 %
		}%
	}%
}
\DeclareRobustCommand\ddots{%
	\mathinner{%
		\mathpalette\@ddots{}%
		\mkern\thinmuskip
	}%
}
\newcommand*{\@ddots}[2]{%
	\sbox0{$#1\cdotp\cdotp\cdotp\m@th$}%
	\sbox2{$#1.\m@th$}%
	\vbox{%
		\dimen@=\wd0 %
		\advance\dimen@ -3\ht2 %
		\kern.5\dimen@
		\dimen@=\wd2 %
		\advance\dimen@ -\ht2 %
		\dimen2=\wd0 %
		\advance\dimen2 -\dimen@
		\vbox to \dimen2{%
			\offinterlineskip
			\hbox{$#1\mathpunct{.}\m@th$}%
			\vfill
			\hbox{$#1\mathpunct{\kern\wd2}\mathpunct{.}\m@th$}%
			\vfill
			\hbox{$#1\mathpunct{\kern\wd2}\mathpunct{\kern\wd2}\mathpunct{.}\m@th$}%
		}%
	}%
}
\makeatother

\let\oldnl\nl
\newcommand{\nonl}{\renewcommand{\nl}{\let\nl\oldnl}}
\makeatother

\def\br{\mathbb R}

\def\sI{\mathcal{I}}

\def\sl{\mathcal{L}}

\def\O{\mathrm{O}}

\newcommand\eqdef{\mathrel{\overset{\makebox[0pt]{\mbox{\normalfont\tiny def}}}{=}}}

\graphicspath{{figures/}}